# An Approach to Stable Gradient Descent Adaptation of Higher-Order Neural Units

Ivo Bukovsky, Noriyasu Homma, *Members IEEE*

*Abstract*— **Stability evaluation of a weight-update system of higher-order neural units (HONUs) with polynomial aggregation of neural inputs (also known as classes of polynomial neural networks) for adaptation of both feedforward and recurrent HONUs by a gradient descent method is introduced. An essential core of the approach is based on spectral radius of a weight-update system, and it allows stability monitoring and its maintenance at every adaptation step individually. Assuring stability of the weight-update system (at every single adaptation step) naturally results in adaptation stability of the whole neural architecture that adapts to target data. As an aside, the used approach highlights the fact that the weight optimization of HONU is a linear problem, so the proposed approach can be generally extended to any neural architecture that is linear in its adaptable parameters.**

## I. Introduction

Higher-order neural units (HONUs) with a polynomial weighting aggregation of neural inputs are known as a fundamental class of polynomial neural networks (PNNs). We may recall that polynomial feedforward neural networks "are attractive due to the reliable theoretical results for their universal approximation abilities according to the Weierstrass theorem [1] and for their generalization power measured by the Vapnik-Chervonenkis (VC) dimension [2]" (cited from [3] with re-numbered references). Work [4] evaluated the computing ability of several types of HONNs by using pseudo-dimensions and VC dimensions [5] and higher-order neural networks (HONNs) were used as a universal approximator. Basically, both the PNNs and HONNs represent the same style of computation in artificial neural networks where neurons involve polynomials, or the neurons are polynomials themselves, or where synaptic connections between neurons involve higher-order polynomial terms. For some of the early publications covering PNN and HONN, we can refer to the technique of extremely high-order polynomial regression tool [6], then to work [7] presenting strong approximating capabilities for a limited number of product nodes while preserving good generalization (low overfitting), to extensions of principal component analysis for higher-order correlations in [8] and optimal fitting hypersurfaces in [9], polynomial basis function discriminant models in [10], to reduced HONN with preservation of geometric invariants in pattern recognition [11], to demonstration of capabilities of HONN for arbitrary dynamical system approximation [12], and to dynamic weight pruning with multiple learning rates in [13]. For more recent works and applications of HONN and PNN we can refer to [14] and the particular focus on a quadratic neural unit (QNU) using matrix notation with upper triangular weight matrix can be found in [15]–[18] and [19]. Significant and most recent publications devoted to PNN concepts are the works [3], [20] [21] while most recent works that are framed within HONNs can be found in [22], where some modifications of HONU are introduced in order to cope with the curse of dimensionality of HONU for higher polynomial orders. Other interesting earlier-appearing neural network architectures are product neural units [23] and later logarithmic neural networks [24]. Another nonconventional neural units are continuous time-delay dynamic neural units and higher-order time-delay neural units that have adaptable time delays in neural synapses and in state feedbacks of individual neurons as introduced in [25]; a similar fuzzy-network oriented concept appeared in parallel also in [26]. Another work focusing on various types of neural transfer functions can be found in review [27]. The optimization of neural weights of conventional neural units is a nonlinear problem, such as for layered networks with hidden neurons with sigmoid output functions. Then nonlinear approaches for stability evaluation that are based on Lyapunov approach or energetic approaches are commonly adopted. Mostly, those techniques are sophisticated and require significant and time demanding (thus costly) effort from users who are not true experts in the field of neural networks. On the other hand, some newer HONU models were proposed for effective computation, learning and configuration in [4] [18] [28]–[34] but they may still suffer from or do not take care of the local minima problem.

To improve the learning of nonlinear adaptive models with gradient descent based learning, we propose a novel approach

Manuscript received in February, 2015. This work was supported in parts by the specific research funding of Czech Technical University in Prague, by grants No. SGS15/189/OHK2/3T/12 and No. SGS12/177/OHK2/3T/12, and also by JSPS KAKENHI Grants No. 25293258 and No. 26540112.

Ivo Bukovsky is with CTU in Prague, Dpt. of Inst. and Control Eng., Adaptive Signal Processing & Informatics Computational Centre (ASPICC), FME, 166 07 Prague, Czech Republic (Ivo.Bukovsky@fs.cvut.cz).

Noriyasu Homma is with the Dpt. of Radiological Imaging and Informatics, Tohoku University Graduate School of Medicine, and also with the Intelligent Biomedical System Engineering Lab., Graduate School of Biomedical Engineering, Tohoku University, 980-8575 Sendai, Japan, (homma@ieee.org).

efficient both in simple stability evaluation and in stability maintenance of GD adaptation and that principally avoids the local minima problems for a given training data set due to in-parameter-linearity of HONUs. The proposed approach recalls that the weight optimization of HONUs, nonlinear input-output mapping models, is a linear problem that theoretically implies the existence of only a unique (global) minimum.

As a minor contribution of this paper, the flattened representation of HONU using 1-D long-vector operations is shown, so the need for multidimensional arrays of weights for HONU is avoided. The introduced long-vector-operator approach also simplifies direct weight calculation of static HONU of an arbitrary polynomial order $r$ by the least square method (LSM), i.e., by variations of Wiener-Hopf equation and points to its connotation to Levenberg-Marquardt (L-M) algorithm for HONU.

The main contribution of this paper is that the adaptation stability evaluation is based on the fact that optimization of weights of HONUs is a linear problem; thus, the evaluation of maximum eigenvalue (spectral radius) can be used to assess the stability of the neural weight system. Based on that principle, the nonlinear extension for stability monitoring and maintenance of static HONU as well as recurrent HONU is proposed. This is the novel approach to evaluation and maintenance of stability of GD adapted HONUs. In principle, the derived stability condition enables gradient adaptation of HONUs be stabilized via time-varying learning rates at every sampling moment. We also discuss the effect of data normalization (more precisely of scaling down the data magnitude), and we show the relationship of the scaling factor to the magnitude of learning rate (in respect to GD adaptation stability). Moreover, our achievements might bring novel research directions for HONU when considering the adaptive learning rate modifications of gradient descent as in [35]. In connection to that, adaptable learning rate modifications for HONU are recalled, and the proposed adaptation stability condition is discussed in connotation to them.

The paper is organized as follows. Subsection II.A introduces the flattening operator (a long-vector operator) approach for HONU, and thus it also reveals the linear optimization nature of HONU. Then, the operator approach is used to derive a stability condition of weight-update system of static HONUs in subsection II.B and of recurrent HONUs in II.C, II.D derives the relationship of data normalization with the change of learning rate for GD. Correspondingly, section III experimentally supports the theoretical derivations, and it also discusses possible extensions of adaptive-learning-rate principles of linear filters [35]–[38] to HONU. The derived adaptation stability rule with adaptive multiple learning rates of static HONUs is demonstrated on the example of up to fifth polynomial order HONU for hyperchaotic Chua's time series prediction, and the rule of dynamic HONU is demonstrated on chaotic Mackey-Glass time series prediction. Also, subsection III.D computationally demonstrates the relationship of decreasing magnitude of data with the decreasing of the learning rate. Basically, we adopt the following standard notation of variables: small caps as "$x$" for a scalar, bolt "$\mathbf{x}$" as for a vector, and bolt capital "$\mathbf{X}$" as for a matrix. Lower indexes as in "$x_i$" or "$w_{i,j}$" denotes the position within a vector or an array, and upper $^\mathrm{T}$ is for the transposition. If a discrete time index is necessary to be shown, it comes as "$k$" in round brackets such as $\mathbf{x}(k)$, $y$ denotes neural output, and $y_p$ is for a training target. The meaning of other symbols is given at their first appearance.

## II. INTRODUCED APPROACHES

### A. Operators for HONU

We start with recalling static QNU [15]–[17], [25] as a fundamental second-order HONU or as a fundamental class of PNN as follows

$$y(k) = \sum_{i=0}^{n} \sum_{j=i}^{n} x_i(k) \cdot x_j(k) \cdot w_{i,j}, \text{ where } x_0 = 1, \quad (1)$$

where $y$ is neural output, $n$ is the number of neural inputs and $w$ stands for neural weights. Let $r$ denote a polynomial order of HONU. Adopting the matrix formulation of QNU from [15], the vector of neural inputs (for al HONU) and the weight array (for QNU) is as follows

$$\mathbf{x}(k) = \begin{bmatrix} x_0 = 1 \\ x_1(k) \\ \vdots \\ x_n(k) \end{bmatrix}, \ \mathbf{W} = \begin{bmatrix} w_{0,0} & w_{0,1} & \cdots & w_{0,n} \\ 0 & w_{1,1} & \cdots & w_{1,n} \\ \vdots & \ddots & \ddots & \vdots \\ 0 & \cdots & 0 & w_{n,n} \end{bmatrix}, \quad (2)$$

where $\mathbf{W}$ is a weight matrix (2-D array for QNU) and $\mathbf{x}(k)$ is input vector at time $k$. Next we drop the time indexing $(k)$ when unnecessary, so $\mathbf{x}=\mathbf{x}(k)$ and $x_i=x_i(k)$. Generally, for HONU of order $r>2$ ($r=3$ for cubic polynomial, $r=4$ for 4th order polynomial, …), such that

$$y = \sum_{i=0}^{n} \sum_{j=i}^{n} \cdots \sum_{\kappa=...}^{n} \left( x_i x_j \cdots x_{...} \right) \cdot w_{i,j,...}, \text{ where } x_0 = 1, \quad (3)$$

the weight $\mathbf{W}$ is understood as a higher-dimensional array (3-D, 4-D,…). In Section II, we derive formulation (13) that is a 1-D array alternative to (1) (3) that allows the gradient descent stability condition of HONU to be effectively derived and that allows connotations to adaptive learning rates of linearly aggregated filters as summarized in [36] (Appendix K). Next, it will be useful to introduce long-vector operators $row^r()$ and $col^r()$ for any polynomial order $r$. As $r=2$ in case of QNU, the operators $row^r()$ and $col^r()$ work as follows

$$row^{r=2}(\mathbf{x}) = row^{r=2}(\mathbf{x}^\mathrm{T}) = [x_0^2 \ \ x_0 x_1 \ ... \ x_i x_j \ \ x_n^2], \quad (4)$$

$$col^{r=2}(\mathbf{x}) = col^{r=2}(\mathbf{x}^\mathrm{T}) = [x_0^2 \ \ x_0 x_1 \ ... \ x_i x_j \ \ x_n^2]^\mathrm{T}, \quad (5)$$

and, e.g., for $r=3$, the $row^3(\mathbf{x})$ would be a row vector as follows

$$row^{r=3}(\mathbf{x}) = [\{x_i x_j x_\kappa\}; i = 0...n, \ j = i...n, \kappa = j...n], \quad (6)$$

where operator $row^r()$ generates a row vector of all $r$-order correlated elements of vector $\mathbf{x}$, and regardless the dimensionality of vector $\mathbf{x}$, it is apparent that

$$col^r(\mathbf{x}) = \left(row^r(x)\right)^T. \tag{7}$$

Letting $N$ denote the total number of training input patterns, then an $n \times N$ matrix $\mathbf{X}$ consisting of all instances of the input vector $\mathbf{x}$ that was defined in (2) is defined as follows

$$\mathbf{X} = [\mathbf{x}(k=1)\ \mathbf{x}(k=2)\ \dots\ \mathbf{x}(k=N)], \tag{8}$$

and when applying the operators defined in (4) (5), $\mathbf{X}$ yields

$$col^r(\mathbf{X}) = [col^r(\mathbf{x}(k{=}1))\ col^r(\mathbf{x}(k{=}2))\ \dots\ col^r(\mathbf{x}(k{=}N))], \tag{9}$$

$$row^r(\mathbf{X}) = \begin{bmatrix} row^r(\mathbf{x}(k{=}1)) \\ row^r(\mathbf{x}(k{=}2)) \\ \vdots \\ row^r(\mathbf{x}(k{=}N)) \end{bmatrix} = \left(col^r(\mathbf{X})\right)^T, \tag{10}$$

where the row indexes of $\mathbf{X}$ stand for input variables (first row is a bias $x_0$=1), while its column indexes correspond to the sample index $k$ (i.e. discrete time index). In general, the application of the $col^r(\mathbf{X})$ or $row^r(\mathbf{X})$ operators on the matrix of input patterns $\mathbf{X}$ means their application to individual input vectors $\mathbf{x}(k)$ of matrix $\mathbf{X}$ as shown in (10). Regarding the neural weights, we can benefit also from the long-column-vector operator or the long-row-vector one for multidimensional weight arrays of HONN (PNN). Notice that the weight matrix of QNU in (1) is a 2-dimensional array, and it would become an $r$-dimensional array of neural weights for an $r$-order neural unit. Therefore, we introduce another, yet compatible, functionality of operators $col()$ and $row()$; this time it is the conversion of multidimensional arrays of neural weights into their long-column-vector or long-row-vector representation. Then for a weight matrix $\mathbf{W}$, e.g. for QNU as in (2), the long-vector operators $col()$ and $row()$ work as follows

$$row(\mathbf{W}) = [w_{0,0}\quad w_{0,1}\quad \dots\quad w_{i,j}\quad \dots\quad w_{n,n}],$$
$$col(\mathbf{W}) = \left(row(\mathbf{W})\right)^T. \tag{11}$$

For clarity of further text, we drop the index of polynomial order $r$, and we also drop the use of round brackets in the operators, so the further notation will be simplified as follows

$$\mathbf{colx} = col^r(\mathbf{x}),\ \mathbf{rowx} = row^r(\mathbf{x}),$$
$$\mathbf{colX} = col^r(\mathbf{X}),\ \mathbf{rowX} = row^r(\mathbf{X}), \tag{12}$$
$$\mathbf{colW} = col(\mathbf{W}),\ \mathbf{rowW} = row(\mathbf{W}).$$

Then in general, an individual output of HONU for any given order $r$ can be calculated by vector multiplication with the use of the above introduced operators as follows

$$y = \mathbf{rowx} \cdot \mathbf{colW} = \mathbf{rowW} \cdot \mathbf{colx}, \tag{13}$$

where "$\cdot$" stands for vector or matrix multiplication and the neural output can be calculated for all time instances by matrix multiplication as follows

$$\mathbf{y} = \mathbf{rowX} \cdot \mathbf{colW}\quad \text{or}\quad \mathbf{y}^T = \mathbf{rowW} \cdot \mathbf{colX}, \tag{14}$$

where $\mathbf{y}$ is $(N \times 1)$ vector of neural outputs. Because we substitute $\mathbf{y}$ and $\mathbf{rowW}$ or $\mathbf{colX}^T$ with measured training data, the optimization of weights in $\mathbf{colW}$ (or $\mathbf{rowW}$) clearly represents a linear set of equations to be solved. Further, we recall the weight calculation for the above static HONU (14) using least squares with the introduced operators. Assume $N$ input vectors in matrix $\mathbf{X}$ as defined in (8), where row indexes stand for input variables and column indexes stand for sampled input patterns. Let $\mathbf{y_p} = [\ y_p(1),\ y_p(2),\ \dots,\ y_p(N)]^T$ denote the $(N \times 1)$ vector of targets for input patterns $\mathbf{X}$. For general polynomial order $r$, we express the square error criteria $Q$ between neural outputs and targets using (14) as follows

$$Q = \sum_{k=1}^{N} \left(y_p(k) - y(k)\right)^2$$
$$= (\mathbf{y_p} - \mathbf{rowX} \cdot \mathbf{colW})^T (\mathbf{y_p} - \mathbf{rowX} \cdot \mathbf{colW}). \tag{15}$$

To calculate weights by the classical least square method (LSM), we solve the set of equations $\partial Q / \partial \mathbf{colW} = \mathbf{0}$, where $\mathbf{0}$ is zero vector with its length as the total number of weights, and that represents the set of equations in a simplified notation as

$$(\mathbf{y_p} - \mathbf{rowX} \cdot \mathbf{colW})^T \cdot \frac{\partial (\mathbf{y_p} - \mathbf{rowX} \cdot \mathbf{colW})}{\partial \mathbf{colW}} = \mathbf{0}, \tag{16}$$

(16) can then be rewritten in a matrix way as

$$(\mathbf{y_p}^T - \mathbf{rowW} \cdot \mathbf{colX}) \cdot (-\mathbf{rowX}) =$$
$$-\mathbf{y_p}^T \cdot \mathbf{rowX} + \mathbf{rowW} \cdot \mathbf{colX} \cdot \mathbf{rowX} = \mathbf{0}. \tag{17}$$

Thus we arrive to the LSM calculation of weights by a variation of the Wiener-Hopf equation for HONU and for arbitrarily polynomial order $r$ in a long-row-vector form as

$$\mathbf{rowW} = \mathbf{y_p}^T \cdot \mathbf{rowX} \cdot \left(\mathbf{colX} \cdot \mathbf{rowX}\right)^{-1}, \tag{18}$$

or alternatively in a long-column-vector as

$$\mathbf{colW} = \left(\mathbf{colX} \cdot \mathbf{rowX}\right)^{-1} \cdot \mathbf{colX} \cdot \mathbf{y_p}. \tag{19}$$

The above formulas for direct calculation of weights by least square method (LSM) imply the existence of a unique solution, i.e., a unique global minimum. Of course, to acquire and select the optimum training data with enough of nonlinearly independent training patterns that would result in correct calculation of weights by LSM is another issue. Then it is practical to notice that we may comfortably derive the Levenberg-Marquardt (L-M) weight-update algorithm for static HONU, e.g. in its simplest form, as follows

$$\Delta \mathbf{colW} = \left(\mathbf{colX} \cdot \mathbf{rowX} + \frac{1}{\mu} \cdot \mathbf{I}\right)^{-1} \cdot \mathbf{colX} \cdot \mathbf{e}, \tag{20}$$

where $\mathbf{e}$ is the standard vector (or matrix) of neural output errors, $\mu$ is learning rate (smaller $\mu$ results in finer weight updates), and $\mathbf{rowX} = \mathbf{colX}^T$ already represents the Jacobian matrix. For automated retraining techniques, it can help to estimate or to try to calculate the weights by LSM (18) or (19) and then to apply L-M algorithm. However, we focus on the stability of gradient descent learning for static and recurrent HONU further in this paper. In this subsection we defined two operators $row()$ and $col()$ for neural architectures with higher-order polynomial aggregation of neural inputs. The functionality of the operators slightly differs when applied to the vector of neural inputs or to the matrix (or multidimensional array) of neural weights. We then used these operators to derive the neural weights using the least square method and thus the existence of single (global) minimum of weight optimization of HONU and the clear relationship of the operators to L-M algorithm was shown. Next, we will use the introduced operators for HONU for evaluation and maintenance of stability of weight-update system at every gradient-descent adaptation step of both static as well as recurrent HONUs.

### B. Weight-Update Stability of Static HONU

The operator approach introduced above can be used for stability evaluation and stability maintenance of weight updates for both static HONU updated by the gradient descent and for recurrent HONU updated by its recurrent version also known as RTRL [39]. We derive the approach for stability evaluation for static HONU in this subsection first. The output of static HONU at discrete time samples $k$ is given in (13). The weight-update system by fundamental gradient descent learning rule for update of all the weights of HONU at sampling time $k$ may be given as

$$\mathbf{colW}(k+1) = \mathbf{colW}(k) + \mu \cdot (y_p - y) \cdot \frac{\partial y}{\partial \mathbf{colW}}, \quad (21)$$

where $y_p$ is the target, $y$ is neural output, $\mu$ is the learning rate (scalar), and

$$\frac{\partial y}{\partial \mathbf{colW}} = \left[ \frac{\partial y}{\partial w_{0,0}} \quad \frac{\partial y}{\partial w_{0,1}} \quad \dots \quad \frac{\partial y}{\partial w_{n,n}} \right]^T. \quad (22)$$

When the neural output $y$ is expressed according to (13) and considering (4)(5)(11)(12), then the derivative of neural output with respect to a single general weight of QNU is as follows

$$\frac{\partial y}{\partial w_{ij}} = \frac{\partial (\mathbf{rowx} \cdot \mathbf{colW})}{\partial w_{ij}} = \mathbf{rowx} \cdot \frac{\partial \mathbf{colW}}{\partial w_{ij}} = x_i \cdot x_j, \quad (23)$$

then the neural weight-update system for a weight of static HONU is as

$$w_{ij}(k+1) = w_{ij}(k) + \mu \cdot \left( y_p - \mathbf{rowx} \cdot \mathbf{colW}(k) \right) \cdot x_i \cdot x_j, \quad (24)$$

and considering (11), a column weight update formula for all weights can be expressed as follows

$$\mathbf{colW}(k+1) = \mathbf{colW}(k) + \mu \cdot \left( y_p - \mathbf{rowx} \cdot \mathbf{colW}(k) \right) \cdot \mathbf{colx}. \quad (25)$$

To proceed further, we expand (25) with consideration of proper vector dimensionality as follows

$$\mathbf{colW}(k+1) = \mathbf{colW}(k) + \mu \cdot \mathbf{colx} \cdot \left( y_p - \mathbf{rowx} \cdot \mathbf{colW}(k) \right)$$

$$= \mathbf{colW}(k) + \mu \cdot \mathbf{colx} \cdot y_p - \mu \cdot \mathbf{colx} \cdot \mathbf{rowx} \cdot \mathbf{colW}(k) \quad (26)$$

$$= \mathbf{colW}(k) - \mu \cdot \mathbf{colx} \cdot \mathbf{rowx} \cdot \mathbf{colW}(k) + \mu \cdot \mathbf{colx} \cdot y_p.$$

Then we separate the parts of the weight-update rule as

$$\mathbf{colW}(k+1) = \left( \mathbf{I} - \mu \cdot \mathbf{colx} \cdot \mathbf{rowx} \right) \cdot \mathbf{colW}(k) + \mu \cdot \mathbf{colx} \cdot y_p \quad (27)$$

where $\mathbf{I}$ is $(n_w \times n_w)$ identity matrix, where $n_w$ is the total number of all weights (also the number of rows of $\mathbf{colW}(k)$). Let's denote the long vector multiplication term as follows

$$\mathbf{S} = \mathbf{colx} \cdot \mathbf{rowx}. \quad (28)$$

Considering that $\mathbf{colx} = \mathbf{rowx}^T$ are external inputs and that $y_p$ is the training target, we clearly see from (27) that the stability condition of the weight-update system of static HONU, as of a linear discrete-time system, is at each time $k$ as follows

$$\rho(\mathbf{I} - \mu \cdot \mathbf{S}) \le 1. \quad (29)$$

where $\rho(.)$ is spectral radius, and $\mathbf{I}$ is an identity matrix of diagonal length equal to the number of neural weights.

To improve the adaptation stability of static HONU, we can update the learning rate $\mu$ and observe its impact on the spectral radius $\rho(.)$. Naturally and instead of single $\mu$, we can introduce time varying individual learning rates for each weight via diagonal matrix $\mathbf{M}$ as

$$\mathbf{M} = \mathbf{M}(k) = diag \left( \mu_{0,0}(k) \ \mu_{0,1}(k) \ \dots \ \mu_{n,n}(k) \right), \quad (30)$$

so the weight-update system becomes

$$\mathbf{colW}(k+1) = \left( \mathbf{I} - \mathbf{M}(k) \cdot \mathbf{S} \right) \cdot \mathbf{colW}(k) + \mathbf{M}(k) \cdot \mathbf{colx} \cdot y_p, \quad (31)$$

and the learning rates on diagonal of $\mathbf{M}$ are ordered accordingly such as the weights in $\mathbf{colW}$ or $\mathbf{rowW}$. The stability of the weight-update system of static HONU at every adaptation step is then classically resulting from (31) as

$$\rho(\mathbf{I} - \mathbf{M}(k) \cdot \mathbf{S}) \le 1, \quad (32)$$

where $\mathbf{S}$ is defined in (28) and the time-indexed learning rate matrix $\mathbf{M} = \mathbf{M}(k)$ indicates that we can stabilize the adaption via time varying learning rates, so (32) is a starting point for developing novel adaptive learning rate algorithms for HONU, e.g., starting with inspiration from works [35], [36] this time for HONU (and other nonlinear models that are linear in their parameters). Also, the condition (32) explains why the normalization of input data affects the gradient descent adaptation stability because high magnitude input data results in large $\mathbf{S}$ (defined in (28)) and that requires small learning rates to approach condition (32) (see II.D).

### C. Weight-Update Stability of Recurrent HONU

Recurrent HONU feeds its step delayed neural output back to its input. The individual weight update of recurrent HONU by fundamental gradient descent (RTRL) can then be given using the above introduced operators and for any polynomial order as follows

$$w_{i,j,\dots}(k+1) = w_{i,j,\dots}(k) +$$

$$+ \mu \cdot \left( y_p(k+n_s) - \mathbf{rowx} \cdot \mathbf{colW}(k) \right) \cdot \frac{\partial y(k+n_s)}{\partial w_{i,j,\dots}}, \quad (33)$$

where $n_s$ is the discrete prediction interval, and the individual derivatives of neural output are for recurrent HONU as follows

$$\frac{\partial y(k+n_s)}{\partial w_{i,j}} = \left( \frac{\partial \mathbf{rowx}(k)}{\partial w_{i,j}} \mathbf{colW}(k) + \mathbf{rowx}(k) \frac{\partial \mathbf{colW}(k)}{\partial w_{i,j}} \right), \quad (34)$$

where weight indexing is shown as if for QNU, and here $\partial \mathbf{rowx}(k) / \partial w_{i,j} \ne \mathbf{0}$ (contrary to static HONU, see (23)) because the neural input $\mathbf{x}$ of recurrent architecture is concatenated with delayed neural outputs, and it can be expressed for all derivatives of neural output in a long-column vector (considering (22) and (23)) as

$$\frac{\partial y(k+n_s)}{\partial \mathbf{colW}} = \left( \frac{\partial \mathbf{rowx}(k)}{\partial \mathbf{colW}} \mathbf{colW}(k) + \mathbf{colx}(k) \right), \quad (35)$$

where when considering (4) for QNU

$$\frac{\partial \mathbf{rowx}}{\partial \mathbf{colW}} = \mathbf{J} = \begin{bmatrix} \dfrac{\partial \mathbf{rowx}}{\partial w_{0,0}} \\ \dfrac{\partial \mathbf{rowx}}{\partial w_{0,1}} \\ \vdots \\ \dfrac{\partial \mathbf{rowx}}{\partial w_{n,n}} \end{bmatrix} = \begin{bmatrix} \dfrac{\partial x_0{}^2}{\partial w_{0,0}} & \dfrac{\partial (x_0 x_1)}{\partial w_{0,0}} & \cdots & \dfrac{\partial x_n{}^2}{\partial w_{0,0}} \\ \dfrac{\partial x_0{}^2}{\partial w_{0,1}} & \dfrac{\partial (x_0 x_1)}{\partial w_{0,1}} & \cdots & \dfrac{\partial x_n{}^2}{\partial w_{0,1}} \\ \vdots & & & \\ \dfrac{\partial x_0{}^2}{\partial w_{n,n}} & \dfrac{\partial (x_0 x_1)}{\partial w_{n,n}} & \cdots & \dfrac{\partial x_n{}^2}{\partial w_{n,n}} \end{bmatrix}, \quad (36)$$

where $\mathbf{J}$ represents the recurrently calculated Jacobian matrix with dimensions $n_w \times n_w$, where $n_w$ is the total number of weights, which is also equal to the number of elements of $\mathbf{rowx}$ or $\mathbf{colW}$. Let us denote $J_{\zeta,\eta}$ the element of Jacobian $\mathbf{J}$ in $\zeta^{\text{th}}$ row and $\eta^{\text{th}}$ column. In case of QNU ($r=2$), $J_{\zeta\eta}$ corresponds to partial derivative of $q^{\text{th}}$ element of vector $\mathbf{rowx}$ that correspond to the second-order polynomial correlation of $i^{\text{th}}$ and $j^{\text{th}}$ neural input $\mathbf{x}$, and the neural output partial derivative by a single weight can be calculated as

$$\frac{\partial y(k+n_s)}{\partial w_{i,j}} = \mathbf{J}_{\zeta,:} \cdot \mathbf{colW}(k) + x_i(k) x_j(k), \quad (37)$$

where $\mathbf{J}_{\zeta,:}$ is the $\zeta^{\text{th}}$ row of Jacobian $\mathbf{J}$ that corresponds to the position of weight $w_{i,j}$ in $\mathbf{colW}$ (and also $\mathbf{rowW}$, see (11)) and it is evaluated as

$$\mathbf{J}_{\zeta,:} = \frac{\partial \mathbf{rowx}}{\partial w_{i,j}} = \begin{bmatrix} \dfrac{\partial (x_0{}^2)}{\partial w_{i,j}} & \dfrac{\partial (x_0 x_1)}{\partial w_{i,j}} & \cdots & \dfrac{\partial (x_n{}^2)}{\partial w_{i,j}} \end{bmatrix}, \quad (38)$$

where its each element is for QNU as follows

$$J_{\zeta,\eta} = \frac{\partial x_\tau(k)}{\partial w_{i,j}} x_\upsilon(k) + x_\tau(k) \frac{\partial x_\upsilon(k)}{\partial w_{i,j}}. \quad (39)$$

Obviously in RTRL, $\partial x_\tau(k)/\partial w_{i,j}$ are calculated recurrently if $x_\tau$ corresponds to the tapped delayed feedback of neural output, and $\partial x_\tau(k)/\partial w_{i,j} = 0$ if $x_\tau$ corresponds to the external input or to a bias $x_0 = 1$. According to (33)-(36) and with correct left-side matrix multiplication, we arrive to RTRL update rule for recurrent HONU of general polynomial order $r$ that considers matrix dimensions for multiplications as

$$\mathbf{colW}(k+1) = \mu \cdot \frac{\partial y(k+n_s)}{\partial \mathbf{colW}} \cdot \left( y_p(k+n_s) - \mathbf{rowx}(k) \cdot \mathbf{colW}(k) \right) \quad (40)$$

$$= (\mathbf{I} + \mu \cdot \mathbf{R} - \mu \cdot \mathbf{S}) \cdot \mathbf{colW}(k) + \mu \cdot \mathbf{colx}(k) \cdot y_p(k+n_s)$$

where the detailed derivation is shown in appendix and where

$$\mathbf{R} = \mathbf{J} \cdot y_p(k+n_s) - \mathbf{J} \cdot \mathbf{colW}(k) \cdot \mathbf{rowx}(k). \quad (41)$$

Again we can introduce a diagonal matrix of learning rates $\mathbf{M}$ instead of a single $\mu$ and separate the parts of the update recurrent system as follows

$$\mathbf{colW}(k+1) = \big(\mathbf{I} + \mathbf{M} \cdot (\mathbf{R} - \mathbf{S})\big) \cdot \mathbf{colW}(k) + \mathbf{M} \cdot \mathbf{colx}(k) \cdot y_p(k+n_s). \quad (42)$$

Then the stability condition for adaptation of recurrent HONU by RTRL technique via evaluation of spectral radius $\rho()$ is as follows

$$\rho\big( \mathbf{I} + \mathbf{M}(k) \cdot (\mathbf{R} - \mathbf{S}) \big) \leq 1 \quad (43)$$

where $\mathbf{S}$ is defined in (28) and the time indexing of the learning rate matrix $\mathbf{M}(k)$ indicates the time variability of individual learning rates. The condition (43) allows us to evaluate and maintain the stability of the update weight system of recurrent HONU at every sampling time $k$. Also, resetting Jacobian $\mathbf{J}$ (36) to zero may be occasionally considered and that results in eliminating term $\mathbf{R}$. Note that when resetting the Jacobian to zero matrix $\mathbf{J} = \mathbf{0}$, the condition for stability of the weight-update system of recurrent HONU (43) yields the stability condition of static HONU (32), and thus the stability of neural weights becomes independent from the actual weights themselves. The proper investigation of the $\mathbf{J}$-reset effect to learning of recurrent HONU and the very rigorous development and analysis of sophisticated techniques for adaptive learning rates, such as based on works [35]–[38] exceeds the limits for this paper; however, the operator approach and the stability conditions of HONU allows us to propose most straightforward connotations to adaptive learning rates techniques for HONU and we introduce them in subsections III.B and III.C.

### D. Data Normalization vs. Learning Rate

The effect of normalization of learning rate can be viewed as the alternative to normalization of magnitude of training data. Let $\alpha$ denote the scaling factor of input data as follows

$$\alpha(\mathbf{x}) = \begin{bmatrix} x_0 = 1 & \alpha \cdot x_1 & \cdots & \alpha \cdot x_n \end{bmatrix}^T; \quad \alpha > 0. \quad (44)$$

For the example of static QNU (HONU with $r=2$) it yields

$$row^{(r=2)}\big(\alpha(\mathbf{x})\big) = \alpha(\mathbf{rowx})$$
$$= [1 \quad \alpha \cdot x_1 \quad \cdots \quad \alpha^{sign(i)+sign(j)} \cdot x_i \cdot x_j \quad \cdots \quad \alpha^2 \cdot x_n{}^2] \quad (45)$$

and for a general order of HONU

$$row^{(r)}\big(\alpha(\mathbf{x})\big) = \begin{bmatrix} 1 & \left\{ \alpha^{sign(i)+sign(j)+\cdots} \cdot x_i \cdot x_j \cdots \right\} \end{bmatrix}. \quad (46)$$

Then it is apparent from the stability condition of static HONU with involvement of scaling factor $\alpha$ as

$$\rho\big( \mathbf{I} - \mu \cdot \alpha(\mathbf{colx}) \cdot \alpha(\mathbf{rowx}) \big) \leq 1, \quad (47)$$

that variation of input data magnitudes and scaling data with $\alpha$ can have up-to *2r-power* stronger influence to adaptation stability than decreasing the learning rate, and it can be stated as

$$\alpha \sim \mu^{2r}; \quad \text{where} \quad 0 < \mu \leq \alpha \leq 1. \quad (48)$$

Experimental comparison of scaling the learning rate versus training data is shown in section III.D.

## III. EXPERIMENTS AND EXTENSIONS

### A. Static HONU (r=3)

In this subsection, we present the results achieved with the proposed operator approach for weight calculation of static HONU by the least square method as derived in subsection II.A. The results support the existence of a unique minimum due to polynomial nonlinearity of HONU. As a benchmark we chose a variation of famous system from [40] as

$$y_{true}(k+1) = \frac{y_{true}(k)}{1 + y_{true}(k)^2} + u(k)^3, \qquad (49)$$

where the training output patterns $y_p$ were obtained as the true values $y_{true}$ with additive measurement noise as

$$y_p(k) = y_{true}(k) + \mathcal{E}(k), \qquad (50)$$

where $u$ and $\varepsilon$ are white noise signals, independent of each other, with unit variances and zero means and signal to noise ratio was calculated commonly as

$$SNR = 10 \cdot \log_{10}\left( \frac{E[y_{true}^2]}{E[\varepsilon^2]} \right) \quad [dB]. \qquad (51)$$

To use the static HONU as a one-step predictor of time series (49) with noise is a suitable task, and weights can be found directly by the least square method even with high noise training data of SNR = 4.83 [dB]. The HONU was trained for first 300 samples and tested on next 700 samples (Fig. 1). Mean absolute error (MAE) of neural output and true signal was 0.43 while the MAE of neural output and noisy training data was 1.691 demonstrating that even if the noise of training data was high, HONU (especially and naturally best for $r=3$) learns the governing laws, approximates the original signal, and tends to reject the noise as seen in Fig. 1.

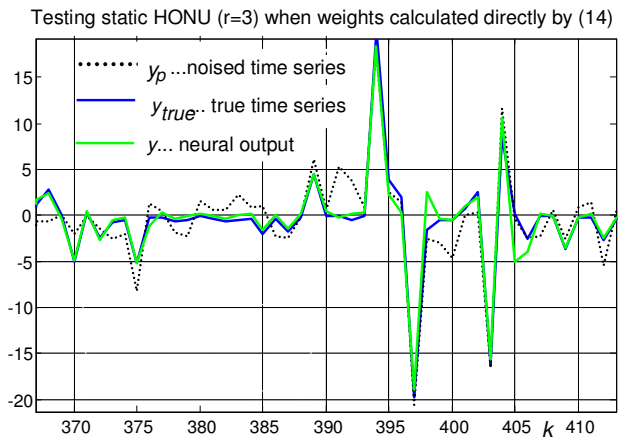

Fig. 1: Testing the static HONU with $r=3$ on benchmark (49) with SNR=4.8 [dB]; apparently HONU extracted the governing laws rather than the noisy training signal (weights found by least squares (19), training data 300 samples, testing data next 700 samples, mean(abs($y_p$-y))=1.6911, mean(abs($y_{true}$-y))=0.4339.

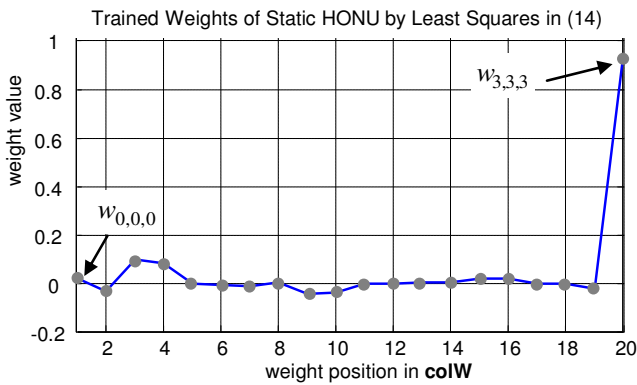

Fig. 2: Twenty weights of static HONU of $r=3$ were calculated by least squares for tested benchmark (49) (Fig. 1); the significant weights are related to $y_p(k)$ and $u(k)$ and to their corresponding multiplicative terms.

Fig. 2 then shows the neural weights directly calculated by the least squares method, and the weights correspond to relevant polynomial terms containing $y(k)$ and $u(k)$ (eg. $w_{3,3,3}$, see (49)), while terms with $y(k-1)$ are suppressed by neural weights that resulted very near to zero (Fig. 2). The results demonstrated the functionality of the proposed operators for LSM weight calculation from the above subsection II.A of static HONU of polynomial order $r=3$. The good quality of nonlinear approximation and the capability of HONU to extract major governing laws from noised training data were also shown. In the next experimental part, we demonstrate the main contribution of this paper, i.e., the results of stability evaluation of weight updates by the gradient descent (GD) method for the recurrent HONU.

This subsection demonstrated good approximation capability of HONU and good extraction of the governing laws even by direct calculation with least squares method (18) or (19), i.e., variations of Wiener-Hopf equations,), which implies the principal existence of single (unique) minima of HONU for a given training data set because HONU are nonlinear models but they are linear in parameters. Next we draw extensions for adaptive learning rates of static and later of recurrent HONUs.

### B. Adaptive Learning Rate of Static HONU

In this subsection, we introduce the connotations of static HONUs to adaptive learning rate techniques that are well known in literature for adaptive filters with linear aggregation of neural inputs [35], [36], [41]–[44].

#### Single Learning Rate

The in-parameter-linearity of HONUs allows us to draw a parallel between HONUs and linearly aggregated (FIR) filters for the learning rate adaptive techniques; i.e., by simple comparison of the long-vector operator notation of HONUs (13) with linearly aggregated FIR filters form (e.g.[36], p.279), we see that $\mathbf{rowx}(k)$ plays the role of $\mathbf{x}(k)$ and the $\mathbf{colW}(k)$ plays the role of $\mathbf{w}(k)$. With those substitutions, we can adapt the learning rate by the classical normalized least mean square (NLMS) algorithm [41], so the adaptive learning rate $\eta$ for static HONUs yields

$$\eta(k) = \frac{\mu}{\|\mathbf{rowx}(k)\|_2^2 + \varepsilon} \qquad (52)$$

were $\varepsilon$ is the regularization term for zero–close input vector. Concluding our experience, we may recommend to use also the square of the Euclidean norm, so (52) yields

$$\eta(k) = \frac{\mu}{\mathbf{rowx}(k) \cdot \mathbf{colx}(k) + \varepsilon}, \qquad (53)$$

and that displayed improved stability and faster convergence in our experiments. The straight explanation for this is as follows. Contrary to (52), the squared-norm normalization in (48) more aggressively contributes to stability condition (29) by suppressing learning rates when the norm of neural inputs exceeds unit, i.e. $\|\mathbf{rowx}(k)\|>1$ that explains better stability (higher learning rate can be used) than with (52). On the other hand, the squared-norm in (48) is naturally more progressive for accelerating adaptation when $\|.\|<1$ that explains faster convergence of GD when normalized with (53). If we introduce $\mathbf{A} = \mathbf{I} - \mu \cdot \mathbf{S}$ according to (29), we can normalize the unique learning rate $\mu$ using the Frobenius norm of $\mathbf{A}$ because it reflects the deviations of spectral radius from a unit and thus it reflects the weight-update stability, so (52) yields

$$\eta(k) = \frac{\mu}{\|\mathbf{A}\|_2^2 + \varepsilon}. \tag{54}$$

Performance comparison of GD with (52)–(56) and other algorithms for static HONU is demonstrated in Fig. 3. Furthemore and similarly to substitution as made in (52) for static HONUs, we may also implement the Benveniste's learning rate updates based on [42], algorithm for Farhang and Ang [43], Mathews' algorithm [44], and generalized normalized gradient descent algorithm (GNGD) of Mandic [35] (as summarized in [36] (Appendix K). We recently showed these extensions for HONU and compared their performance for chaotic time series in [45]. They are recalled in Tab. 1.

### *Multiple Learning Rates*

As indicated in subsection II.B and as it also resulted from our experiments, it appeared more efficient when we used individual learning rates for each weight and normalize them individually. We propose the following algorithm. When we redefine $\mathbf{A} = (\mathbf{I} - \mathbf{M}(k) \cdot \mathbf{S})$ according to (32) for multiple learning rates in $\mathbf{M}(k)$, we can contribute to the stability by normalization of individual learning rates in $\mathbf{M}(k)$ by Euclidean norm of corresponding rows in $\mathbf{A}$, because individual learning rates in $\mathbf{M}(k)$ multiply only corresponding rows in $\mathbf{S}$ and thus they affect only the corresponding rows in $\mathbf{A}$. Therefore the adaptive learning rate can be normalized as

$$\eta_q(k) = \frac{\mu_q}{\|\mathbf{A}_{q,:}\|_2^2 + \varepsilon}, \tag{55}$$

where $q$ is a position index of a weight in **colW** and it also indexes the corresponding learning rate in diagonal matrix of learning rates $\mathbf{M}$, and $\mathbf{A}_{q,:}$ is $q^{\text{th}}$ row of matrix $\mathbf{A}$. Again, we may use the squared Euclidean norm of the rows, so (55) yields

$$\eta_q(k) = \frac{\mu_q}{\mathbf{A}_{q,:} \cdot (\mathbf{A}_{q,:})^T + \varepsilon}, \tag{56}$$

Again, the modification of normalization algorithm (56) performed faster convergence than (52)–(55) (provided manually optimized $\varepsilon$ that was easy to be found $\varepsilon=1$). Again, the squared norm in (56) is more aggressive than the norm itself, and the individual learning rates are normalized so they equally contribute to stability. Practically, we have not found too significant difference in performance between (52)–(56), because tuning of $\varepsilon$ plays its role that we do not focus in this paper. Anyhow, we found the above adaptive learning rate techniques (52)–(56) very useful for static HONU, and they are in clear connotation to the weight update stability (29). Yet it appeared in experiments that the normalization for static HONU by (56) is more efficient (faster and maintaining spectral radius closest to 1) than the above mentioned options (52)–(55). To us, it practically appeared that the normalizing approaches (52)–(56) are for static HONU superior both in speed and in maintaining convergence for long adaptation runs (or for many epochs), while the gradient based learning rate adaptive techniques, adopted for HONU, tend to require an early stopping and that is a well-known issue.

**Tab. 1: Extensions of adaptive learning rates for HONU [45]; the adaptive learning rate then still can be used in stability conditions (29)**

| | |
|---|---|
| **Based on Normalized Least Mean Squares (or Normalized GD)** | |
| NLMS [41] | $\Delta\mathbf{w} = \dfrac{\mu}{\varepsilon + \|\mathbf{colx}(k-p)\|_2^2} \cdot e(k) \cdot \mathbf{colx}(k-p)^T$ |
| GNGD [35] | $\varepsilon(k+1) = \varepsilon(k) - \beta \cdot \mu \cdot \dfrac{e(k)e(k-1)\mathbf{colx}(k-p)^T \cdot \mathbf{colx}(k-p-1)}{\left(\|\mathbf{colx}(k-p-1)\|_2^2 + \varepsilon(k)\right)^2}$ |
| RR–NLMS [37] | $\varepsilon(k+1) = \max\left[\varepsilon_{\min},\right.$ $\varepsilon(k) - \beta \cdot sign\left(e(k) \cdot e(k-1) \cdot \mathbf{colx}(k-p)^T \cdot \mathbf{colx}(k-p-1)\right)\left.\right]$ |
| **Based on Performance Index Derivative** $\mu(k+1) = \mu(k) + \beta \cdot e(k) \cdot \boldsymbol{\gamma}(k) \cdot \mathbf{colx}(k-p)$ | |
| Benveniste's [42] | $\boldsymbol{\gamma}(k+1) = \left[\mathbf{I} - \mu(k-1)\mathbf{colx}(k-p-1) \cdot \mathbf{colx}(k-p-1)^T\right]\boldsymbol{\gamma}(k-1)$ $+ e(k-1) \cdot \mathbf{colx}(k-p-1)$ |
| Farhang's & Ang's [43] | $\boldsymbol{\gamma}(k) = \eta \cdot \boldsymbol{\gamma}(k-1) + e(k-1) \cdot \mathbf{colx}(k-p-1); \ \eta \in \langle 0 \ , \ 1 \rangle$ |
| Mathew's [44] | $\mu(k+1) = \mu(k) + \beta \cdot e(k) \cdot e(k-1) \cdot \mathbf{colx}(k-p)^T \cdot \mathbf{colx}(k-p-1)$ $(\boldsymbol{\gamma}(k) = 1)$ |

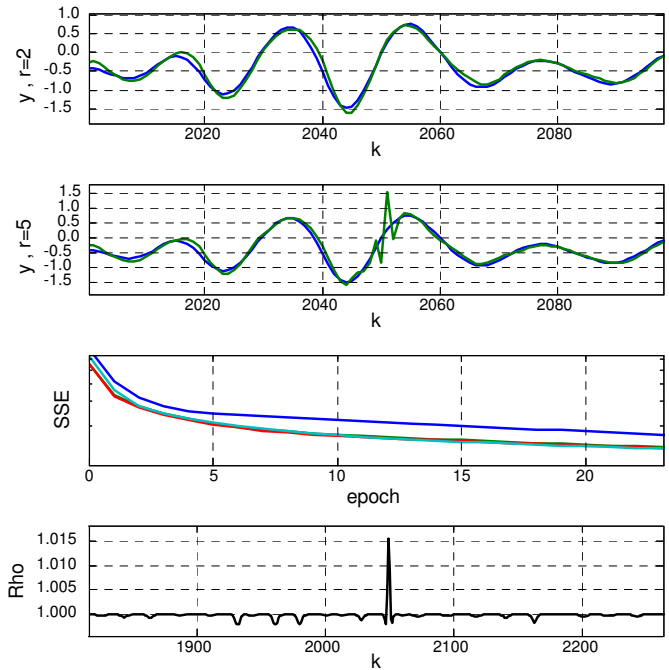

Fig. 3: GD training of HONUs $r=2$ (top axis) , $r=5$ (second axis), sum of square errors (third axes) during training epochs of HONU $r=2,3,4,5$; and the real-time estimation of the spectral radius increase (32) by Frobenius norm in the last epoch of training (bottom axis) of $5^{\text{th}}$ order HONU.

An explanation can be that the normalizing approaches contribute to the derived stability conditions of weight updates (29) or (43), while the gradient adaptive learning rate techniques do not consider the stability maintenance so straightforwardly. The performance of the learning with adaptive learning rates (56) and showing the stability condition (32) for static HONUs of order $r=2,3,4,5$ for prediction of hyperchaotic Chua's time series [46]–[48] is shown in Fig. 3.

### C. Recurrent HONU

In this part, we demonstrate the validity of the introduced weight-update stability condition of recurrent HONU (43).

**Stability Monitoring**

Let us use recurrent QNU (HONU r=2) for a long-term prediction of the MacKey-Glass time series in chaotic mode [49], [50] that is given as

$$\dot{x}(t) = 0.2\, x(t-\tau) \cdot \left(1 + x(t-\tau)^{10}\right)^{-1} - 0.1 x(t)\,, \qquad (57)$$

where $\tau=17$ and the time series was obtained with 1 sampling per second. Configuration of HONU as a nonlinear recurrent predictor was the prediction time $n_s=11$ steps (seconds) and the input of HONU included bias $x_0=1$ and 10 tapped delayed feedbacks and 7 most recently measured values. Fig. 4 shows the later epoch of stable adaptation of recurrent HONU being trained according to the gradient descent learning rule and using the operator approach as derived in subsection II.C.

The adaptation in Fig. 4 was stable because of a sufficiently small learning rate, and the occasional violations of stability condition (43) in the bottom plot of Fig. 4 spontaneously diminished and have not resulted in instability of recurrent HONU. The example of unstable adaptation of recurrent HONU is given in Fig. 5 where the weight update becomes unstable before $k=700$ and which appears as oscillations of neural output (top plot in Fig. 5) and thus as oscillations of error (middle plot

Fig. 5). Importantly, the stability condition (43) (bottom plot in Fig. 5) became significantly violated before neural output oscillations appeared and this is well apparent from detail in Fig. 6. We can see in bottom plots of Fig. 5 and Fig. 6 that the weight system returns to stability again after $k=717$ (spectral radius returned very close to one, $\rho \approx 1$ (43)). This was maintained by the simplest way that the proposed approach offers, i.e., if the spectral radius (43) exceeded a predefined threshold (here $\rho>1.05$), we decreased the learning rate (here we used a single learning rate for all weights $\mu \leftarrow 0.6 \cdot \mu$ ) and we reset the Jacobian to zeros and recurrently calculated gradients that was shown in (39). As mentioned already, the more advanced stability maintenance can be carried out by introducing individual learning rates (30)(32) for each weight and to optimize their magnitudes with respect to stability condition for static HONU according to (32) and with respect to stability condition for recurrent HONU according to (43) (exploring this deserves further research and exceeds this paper).

### D. Data Normalization vs. Learning Rate

To computationally verify the derivation in subsection II.D, normally distributed zero-mean data of unit standard deviation were used as original input data into vector **x** of various length $n$ for static QNU. The effect of scaling factor $\alpha$ versus the one of the learning rate $\mu$ on adaptation stability (32) and thus confirming the relationship (48) is demonstrated via Fig. 7 to Fig. 10. For data with larger variance of magnitude, Fig. 7–Fig. 10 imply that adaptive $\mu$ requires be adapted within much wider interval, approximately $\mu \in (1\mathrm{E}-4\ ,\ 1)$ , of values rather than when data are normalized (scaled down), approximately as $\mu \in (1\mathrm{E}-1\ ,\ 1)$ .

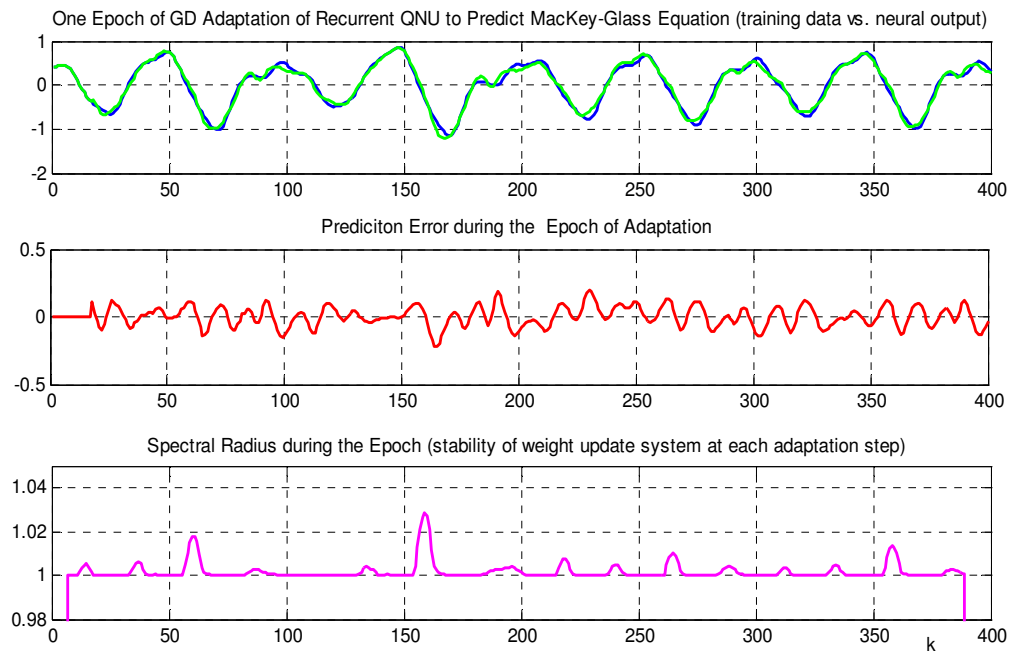

Fig. 4: Stable adaptation of recurrent HONU of $r=2$; the bottom plot monitors the stability (i.e. the spectral radius (43)) of the weight-update system (33).

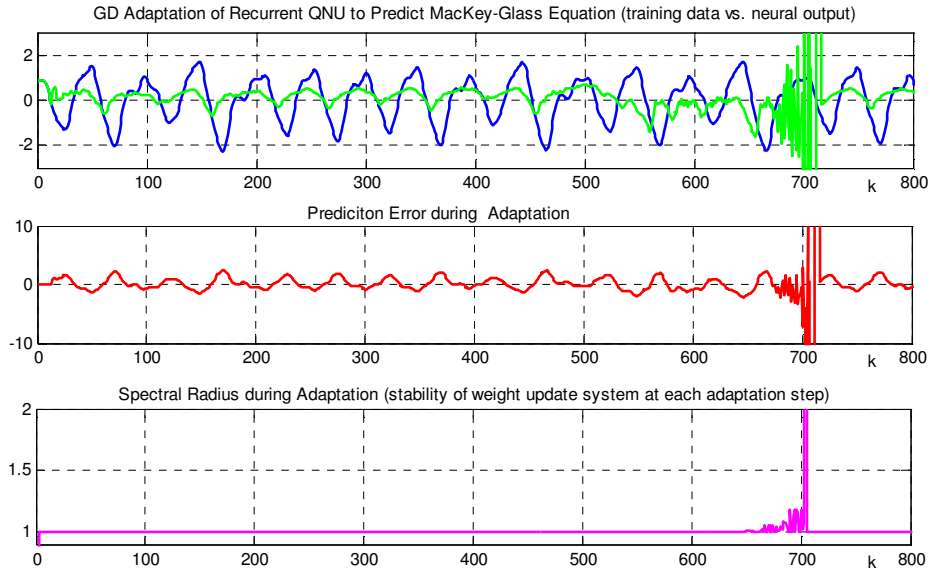

Fig. 5: Unstable adaptation - the bottom plot monitors the stability of weight-update system (33) of recurrent HONU of $r=2$, instability of weights originates at around $k=650$ (see detail in Fig. 6).

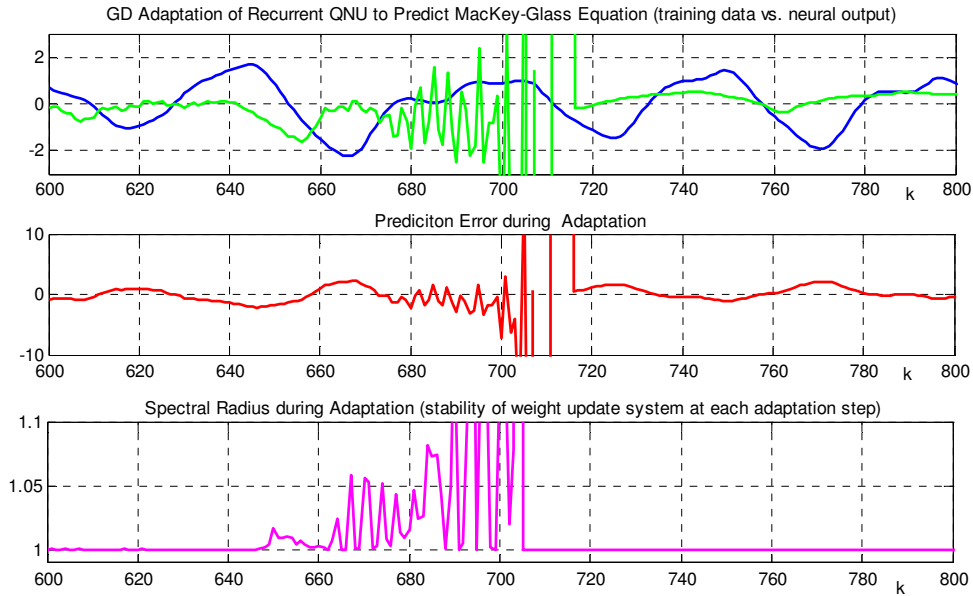

Fig. 6: Unstable adaptation – detail of Fig. 5, the stability condition (43) (bottom plot) became significantly violated well before unusually large oscillations and divergence of neural output.

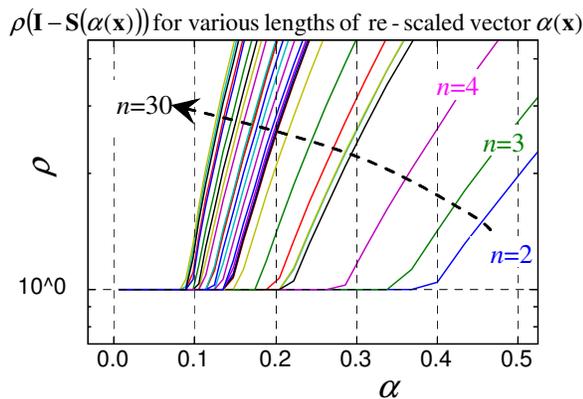

Fig. 7:  Spectral radius $\rho$ of static QNU as the function of both the number of inputs $n$ and the scaling–down factor $\alpha$.

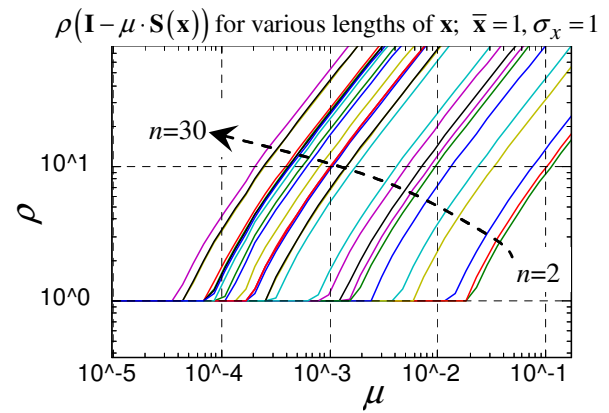

Fig. 8: Spectral radius $\rho$ of static QNU as the function of both the number of inputs $n$ and the learning rate $\mu$.

## IV. DISCUSSION

Besides the good quality of nonlinear approximation, the first example in subsection III.A demonstrated that weights of static HONU can be calculated by the least mean square (LMS) approach using the introduced operators, and it also supports the fact of existence of a single minimum for weight optimization of HONU; the linear nature of the weight optimization task is apparent already from neural output equations of HONU (15) and linear problems have only a single solution. Although the weight optimization by LMS for the benchmark in subsection III.A was a suitable task for static HONU, it is known that adaptation by GD becomes nontrivial task for this benchmark because the weight-update system by GD becomes unstable and requires the control of magnitude of learning rate (several approaches, but not for HONU, to prevent instability and to improve convergence are known [35], [37], [38]).

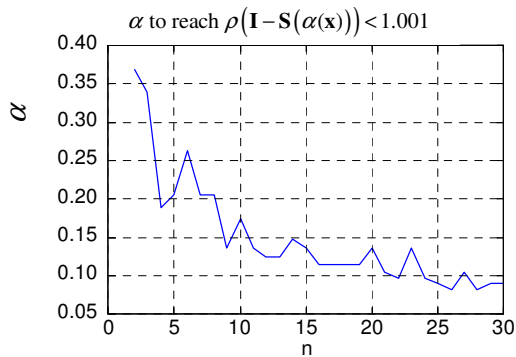

Fig. 9: Computationally estimated data scaling factor $\alpha$ to reach spectral radius $\rho$=1.001; the figure roughly confirms the $2r - power$ relationship to variation of learning rate $\mu$ as shown in (Fig. 10).

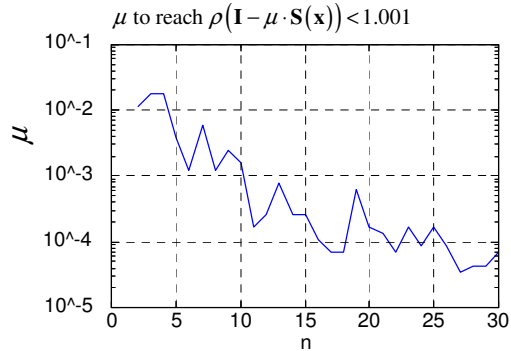

Fig. 10: Computationally estimated learning rate $\mu$ to reach spectral radius $\rho$=1.001; the figure roughly confirms the $2r - power$ relationship to scaling factor $\alpha$ shown in (Fig. 9).

As an aside, it was recalled that optimization of HONU features a unique minimum for weight optimization with a given training data that contains enough of linearly independent training patterns (equal to or more than the number of neural weights) and demonstrated on benchmark data. The proposed approach with HONU is helpful for many identification and control applications. Also the struggle with overfitting can be relieved from local minima issue introduced by the neural architecture itself and thus the effort to reach good generalization of nonlinear model can be focused primarily to proper data processing and to finding appropriate input configuration. We practically observed in [51], [52] that weight convergence of static HONU using L-M algorithm was very rapid and required a very few epoch in comparison to conventional multilayered perceptron networks. Moreover, HONUs that were trained from various initial weights had almost identical outputs compared to various instances of trained MLP networks whose outputs were different for the same input patterns when trained from different initial weights and for the same configurations [51], [52]. This can be reasoned in principle by the above recalled linear nature of the weight optimization of static HONU that implies the existence of a single minimum for a particular input configuration and for a given training data set, while conventional MLP suffers from local minima. The introduced operator approach and online weight-update stability evaluation of a gradient descent method is applicable to any neural architecture that is linear in its parameters if the neural output can be expressed as by (14) where **colX** or **rowX** may also include other nonlinear terms than the multiplicative ones as in case of HONU in this paper.

For the introduced adaptation stability of HONU, we also derived and experimentally showed in subsections II.D and III.D that scaling of the training data by a factor **R** has up-to $2r-power$ stronger effect to adaptation stability (of $r^{th}$ polynomial HONU with up to 30 inputs) than the variation of its learning rate $\mu$.

The requirement for larger interval of $\mu$ implies a possible need for its faster adaptation for un-normalized data, while the adaptation of $\mu$ does not have to be so fast when data are normalized.

As regards the estimation of time complexity in sense of required computational operations per one sample time, the output of $r$-th order HONU with $n$ inputs is calculated as vector multiplication **rowW** · **colx**, where both vectors have length $n_w = \binom{n}{r}$, and because each element of **colx** is made of $r$-th order polynomial terms, the computational complexity of HONU is $O\left(r \cdot \binom{n}{r}\right) = O\left(r \cdot n_w\right)$. For the case of static HONU with the introduced weight update stability, the stability condition (32) requires the Frobenius norm calculation of $n_w \times n_w$ matrix **S**. Thus, the weight update stability of static HONU (32) results in major time complexity of $O\left(n_w^2\right)$. For the case of dynamical HONU, the weight update stability condition (43) requires computation of matrix **R** that involves matrix multiplication of two matrices each of $n_w \times n_w$ elements (41), thus the time complexity estimation can be increased to $O\left(n_w^3\right)$. When a true spectral radius shall be calculated instead of a matrix norm, the time complexity of HONU with weight update stability would approximately increase up to $O\left(n_w^3\right)$ for static HONU and to $O\left(n_w^4\right)$ for recurrent HONU. From the practical point of view and based on our

 

observations; however, it should be mentioned that HONUs can be found useful and efficient esp. for small network problems, i.e. for up to 30 to 50 inputs and many problems can be sufficiently solved with HONU of order $r \le 3$. For such problems, the time complexity of the introduced algorithm shall not be a practical issue with nowadays hardware.

## V. CONCLUSIONS

Using the introduced long-vector notation, the approach to the gradient descent adaptation stability of static and recurrent HONUs of a general polynomial order $r$ was introduced via monitoring of the spectral radius of the weight-update systems at every adaptation step. In experiments, the method was verified as the adaptation instability was detected well before the prediction error divergence became visually clear. Due to in-parameter linearity of HONU, adaptive learning rate techniques for HONU were adopted as known from the linear adaptive filters, and the adaptation stability monitoring was applied to HONUs as well. Also, it was derived and experimentally shown that scaling-down of the training data by a factor $\alpha$ takes up-to *2r-power* stronger influence to adaptation stability rather than the decrease of the learning rate itself. This implies the importance of training data normalization, esp., for adaptive learning rate techniques.

By the presented approaches, HONUs are highlighted as neural architectures that offer adjustable strong nonlinear input-output mapping models with linear optimization nature (thus without local minima issues for a given training data set), and we propose a novel yet comprehensible approach toward stability of the gradient descent weight-update system that can be useful in prediction, control and system monitoring tasks.

## SPECIAL THANKS

Special thanks belongs to Madan M. Gupta from the Intelligent System Research Laboratory of the University of Saskatchewan for founding and forming the author's collaborative research on HONU since 2003 and for long-term support and advices.

The authors also thank to Jiri Fürst from Dpt. of Technical Mathematics, Faculty of Mech. Eng. CTU in Prague, for consultations.

Special thanks belongs to The Matsumae International Foundation (Tokyo, Japan) that funded the first author's cooperation with colleagues at Tohoku University in Japan in 2009, and this cooperation is still vitally continuing.

The authors would like to thank the anonymous reviewers and the Editor in Chief for their insightful and constructive remarks.

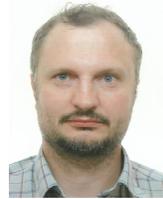

**Ivo Bukovsky** graduated from Czech Technical University (CTU) in Prague where he received his Ph.D. in the field of Control and System Engineering in 2007. Ivo is currently with the Department of Instrumentation and Control Engineering at CTU in Prague, and he is forming the Adaptive Signal Processing and Informatics Computational Centre (ASPICC) within his department. His research interests include higher-order neural networks, adaptive novelty detection, dynamical systems, real-data based modeling, multiscale-analysis approaches, adaptive control and biomedical applications. Ivo was a long-term visiting researcher at the Intelligent Systems Research Lab. at the University of Saskatchewan in Canada (2003), at the Cyberscience Center at Tohoku University in Japan (2009, 2011), and a short-term visiting researcher at the Dpt. of Electrical and Computer Engineering at the University of Manitoba (Witold Kinsner, 2010). In 2016, Ivo is appointed a visiting Associate Professor to Tohoku University.

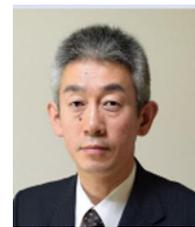

**Noriyasu Homma** (M'99) received a BA, MA, and PhD in electrical and communication engineering from Tohoku University, Japan, in 1990, 1992, and 1995, respectively. From 1995 to 1998, he was a lecturer at the Tohoku University, Japan. He is currently full professor at the Dpt. of Radiological Imaging and Informatics, Tohoku University Graduate School of Medicine, and he is also with the Intelligent Biomedical System Engineering Lab., Graduate School of Biomedical Engineering, both at the Tohoku University. From 2000 to 2001, Noriyasu was a visiting professor at the Intelligent Systems Research Laboratory, University of Saskatchewan, Canada. His current research interests include neural networks, complex and chaotic systems, soft-computing, cognitive sciences, medical systems and brain sciences. He has published over 100 journal and conference papers, edited 1 book, and co-authored 1 book and more than 10 chapters in 10 research books in these fields. He has been serving as PC for many IEEE conferences and as an associated editor of several international journals.

APPENDIX

STABILITY OF THE WEIGHT UPDATE SYSTEM OF RECURRENT HONNU

Starting as for a single weight, the weight update of recurrent HONU begins as

$$w_{i,j,\dots}(k+1) = w_{i,j,\dots}(k) + \mu \cdot \frac{\partial y(k+n_s)}{\partial w_{i,j,\dots}} \cdot \left( y_p(k+n_s) - \mathbf{rowx}(k) \cdot \mathbf{colW}(k) \right) = w_{i,j,\dots}(k) + \mu \cdot \frac{\partial \mathbf{rowx}(k) \cdot \mathbf{colW}(k)}{\partial w_{i,j,\dots}} \cdot \left( y_p(k+n_s) - \mathbf{rowx}(k) \cdot \mathbf{colW}(k) \right)$$

$$= w_{i,j,\dots}(k) + \mu \cdot \left( \frac{\partial \mathbf{rowx}(k)}{\partial w_{i,j,\dots}} \mathbf{colW}(k) + \mathbf{rowx}(k) \frac{\partial \mathbf{colW}(k)}{\partial w_{i,j,\dots}} \right) \cdot \left( y_p(k+n_s) - \mathbf{rowx}(k) \cdot \mathbf{colW}(k) \right)$$

$$= w_{i,j,\dots}(k) + \mu \cdot \left( \frac{\partial \mathbf{rowx}(k)}{\partial w_{i,j,\dots}} \mathbf{colW}(k) + x_i \cdot x_j \cdots \right) \cdot \left( y_p(k+n_s) - \mathbf{rowx}(k) \cdot \mathbf{colW}(k) \right),$$

and with the use of the long-vector operator approach, it can be for all weights as follows

$$\mathbf{colW}(k+1) = \mathbf{colW}(k) + \mu \cdot \left( \frac{\partial \mathbf{rowx}(k)}{\partial \mathbf{colW}} \mathbf{colW}(k) + \mathbf{colx}(k) \right) \cdot \left( y_p(k+n_s) - \mathbf{rowx}(k) \cdot \mathbf{colW}(k) \right).$$

When introducing a simpler notation for Jacobian $\partial \mathbf{rowx}(k) / \partial \mathbf{colW} = \mathbf{J}$

$$\mathbf{colW}(k+1) = \mathbf{colW}(k) + \mu \cdot \left( \mathbf{J} \cdot \mathbf{colW}(k) + \mathbf{colx}(k) \right) \cdot \left( y_p(k+n_s) - \mathbf{rowx}(k) \cdot \mathbf{colW}(k) \right)$$

$$= \mathbf{colW}(k) + \mu \cdot \left( \mathbf{J} \cdot \mathbf{colW}(k) \cdot y_p(k+n_s) - \mathbf{J} \cdot \mathbf{colW}(k) \cdot \mathbf{rowx}(k) \cdot \mathbf{colW}(k) - \mathbf{colx}(k) \cdot \mathbf{rowx}(k) \cdot \mathbf{colW}(k) + \mathbf{colx}(k) \cdot y_p(k+n_s) \right)$$

$$= \mathbf{colW}(k) + \mu \cdot \left( \mathbf{J} \cdot y_p(k+n_s) - \mathbf{J} \cdot \mathbf{colW}(k) \cdot \mathbf{rowx}(k) - \mathbf{colx}(k) \cdot \mathbf{rowx}(k) \right) \cdot \mathbf{colW}(k) + \mu \cdot \mathbf{colx}(k) \cdot y_p(k+n_s),$$

and when having $\mathbf{I}$ for an identity matrix

$$\mathbf{colW}(k+1) = \left( \mathbf{I} + \mu \cdot \left( \mathbf{J} \cdot y_p(k+n_s) - \mathbf{J} \cdot \mathbf{colW}(k) \cdot \mathbf{rowx}(k) - \mathbf{colx}(k) \cdot \mathbf{rowx}(k) \right) \right) \cdot \mathbf{colW}(k) + \mu \cdot \mathbf{colx}(k) \cdot y_p(k+n_s).$$

When introducing auxiliary notation for matrices as

$$\mathbf{R} = \mathbf{J} \cdot y_p(k+n_s) - \mathbf{J} \cdot \mathbf{colW}(k) \cdot \mathbf{rowx}(k)$$

$$\mathbf{S} = \mathbf{colx}(k) \cdot \mathbf{rowx}(k),$$

and for total of $n_w$ weights and with the matrix of time variable learning rates

$$\mathbf{M}(k) = diag \left( \mu_0(k) \quad \mu_1(k) \quad \dots \quad \mu_q(k) \quad \dots \quad \mu_{n_w}(k) \right),$$

the stability condition of recurrent HONU involves both $\mathbf{S}$ and $\mathbf{R}$ is as follows

$$\rho \left( \mathbf{I} + \mathbf{M}(k) \cdot (\mathbf{R} - \mathbf{S}) \right) \leq 1,$$

while for static architecture (or when Jacobian is zero matrix $\mathbf{J} = \mathbf{0}$), the stability condition involves only $\mathbf{S}$:

$$\rho \left( \mathbf{I} - \mathbf{M}(k) \cdot \mathbf{S} \right) \leq 1,$$

where $\rho(.)$ is spectral radius.